\documentclass[pdflatex,sn-basic,iicol]{sn-jnl}

\usepackage{bm}
\usepackage[american]{babel}
\usepackage{booktabs} 
\usepackage{microtype}
\usepackage{graphicx}
\usepackage{subfigure}
\usepackage{booktabs} 
\usepackage{multirow}
\usepackage{hyperref}

\usepackage{amsmath}
\usepackage{amssymb}
\usepackage{amsthm}
\usepackage{mathtools} 
\usepackage{cleveref}

\jyear{2022}%

\theoremstyle{thmstyleone}%
\newtheorem{theorem}{Theorem}
\newtheorem{proposition}{Proposition}
\newtheorem{assumption}{Assumption}
\theoremstyle{thmstyletwo}%
\newtheorem{example}{Example}%

\theoremstyle{thmstylethree}%
\newtheorem{definition}{Definition}%

\begin{document}

\title[ ]{Probabilistic Time Series Forecasts with Autoregressive Transformation Models}


\author*[1,2]{\fnm{David} \sur{R\"ugamer}}\email{david@stat.uni-muenchen.de}

\author[3]{\fnm{Philipp F.M.} \sur{Baumann}}\email{baumann@kof.ethz.ch}

\author[4]{\fnm{Thomas} \sur{Kneib}}\email{tkneib@uni-goettingen.de}

\author[5]{\fnm{Torsten} \sur{Hothorn}}\email{torsten.hothorn@uzh.ch}

\affil*[1]{\orgdiv{Department of Statistics}, \orgname{LMU Munich},  \orgaddress{\city{Munich}, \country{Germany}}}

\affil[2]{\orgdiv{Institute of Statistics}, \orgname{RWTH Aachen}, \orgaddress{\city{Aachen}, \country{Germany}}}

\affil[3]{\orgdiv{KOF Swiss Economic Institute}, \orgname{ETH Zurich}, \orgaddress{\city{Zurich},  \country{Switzerland}}}

\affil[4]{\orgdiv{Chair of Statistics}, \orgname{University of Goettingen}, \orgaddress{\city{Goettingen},  \country{Germany}}}

\affil[5]{\orgdiv{Epidemiology, Biostatistics and Prevention Institute}, \orgname{University of Zurich}, \orgaddress{\city{Zurich},  \country{Switzerland}}}


\abstract{Probabilistic forecasting of time series is an important matter in many applications and research fields. In order to draw conclusions from a probabilistic forecast, we must ensure that the model class used to approximate the true forecasting distribution is expressive enough. Yet, characteristics of the model itself, such as its uncertainty or its feature-outcome relationship are not of lesser importance. This paper proposes Autoregressive Transformation Models (ATMs), a model class inspired by various research directions to unite expressive distributional forecasts using a semi-parametric distribution assumption with an interpretable model specification. We demonstrate the properties of ATMs both theoretically and through empirical evaluation on several simulated and real-world forecasting datasets.}

\keywords{Semi-parametric Models, Conditional Density Estimation, Distributional Regression, Normalizing Flows}



\maketitle

\section{Introduction}

Conditional models describe the conditional distribution $F_{Y\mid x}(y\mid \bm{x})$ of an outcome $Y$ conditional on observed features $\bm{x}$ \citep[see, e.g.,][]{Jordan.2002}. Instead of modeling the complete distribution of $Y\mid \bm{x}$, many approaches focus on modeling a single characteristic of this conditional distribution. Predictive models, for example, often focus on predicting the average outcome value, i.e., the expectation of the conditional distribution. Quantile regression \citep{Koenker.2005}, which is used to model specific quantiles of $Y\mid \bm{x}$, is more flexible in explaining the conditional distribution by allowing (at least theoretically) for arbitrary distribution quantiles. Various other approaches allow for an even richer explanation by, e.g., directly modeling the distribution's density $f_{Y\mid \bm{x}}$ and thus the whole distribution $F_{Y\mid x}(y\mid \bm{x})$. Examples include mixture density networks \citep{Bishop.1994} in machine learning, or, in general, probabilistic modeling approaches such as Gaussian processes or graphical models \citep{Murphy.2012}. 
In statistics and econometrics, similar approaches exist, which can be broadly characterized as distributional regression (DR) approaches \citep{Chernozhukov.2013, Foresi.1995, Ruegamer.2020, Wu.2013}. Many of these approaches can also be regarded as conditional density estimation (CDE) models.

\begin{figure}[ht]
    \centering
    \includegraphics[width=0.95\columnwidth]{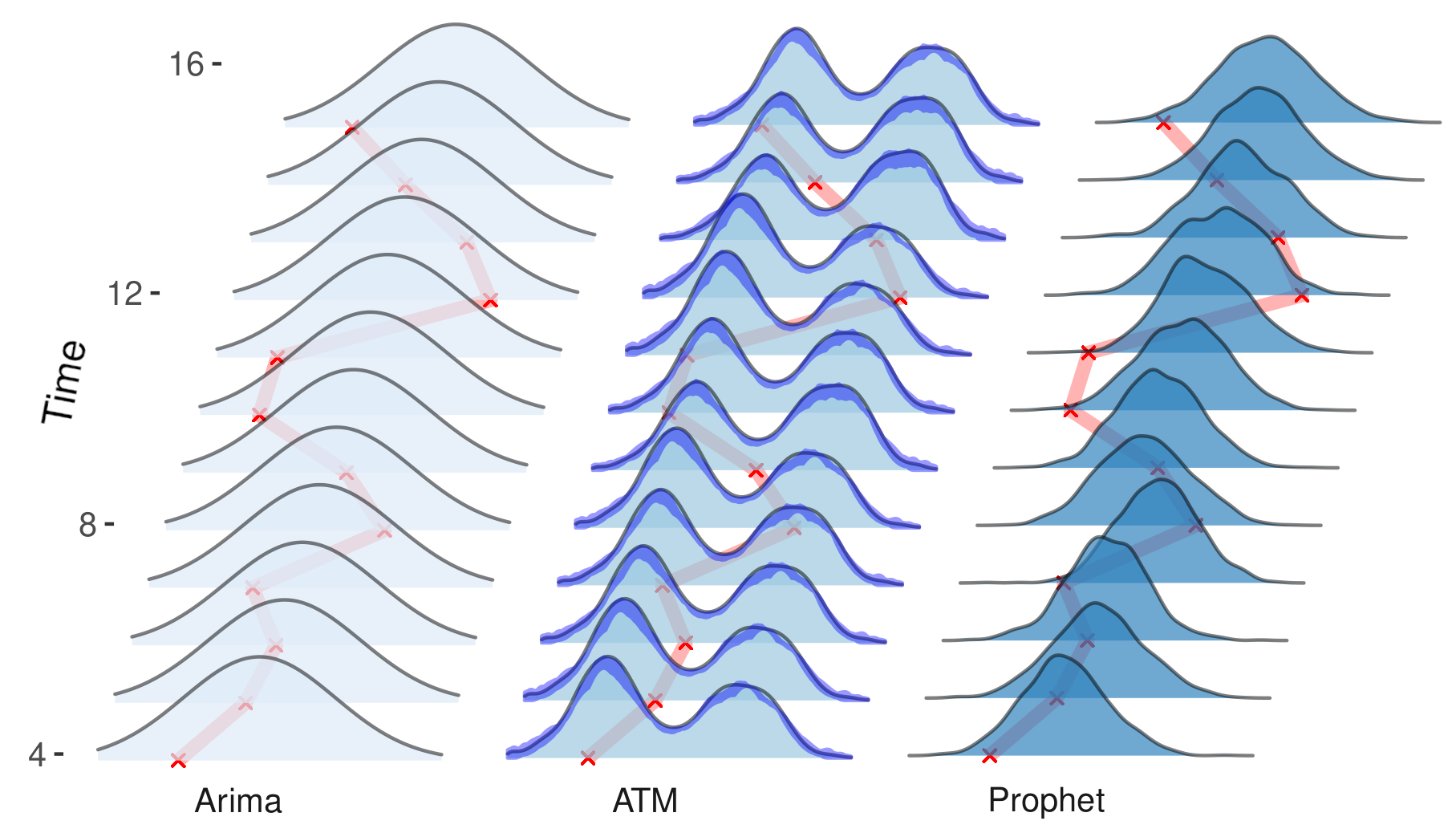}
    \caption{Exemplary comparison of probabilistic forecasting approaches with the proposed method (ATM; with its uncertainty depicted by the darker shaded area) for a given time series (red line). While other methods are not expressive enough and tailored toward a simple unimodal distribution, our approach allows for complex probabilistic forecasts (here a bimodal distribution where the inducing mixture variable is unknown to all methods).}
    \label{fig:motivating}
\end{figure}

Modeling $F_{Y\mid x}(y\mid \bm{x})$ is a challenging task that requires balancing the representational capacity of the model (the expressiveness of the modeled distribution) and its risk for overfitting. While the inductive bias introduced by parametric methods can help to reduce the risk of overfitting and is a basic foundation of many autoregressive models, their expressiveness is potentially limited by this distribution assumption (cf. Figure~\ref{fig:motivating}).

\paragraph{Our contributions}

In this work, we propose a new and general class of semi-parametric autoregressive models for time series analysis called \emph{autoregressive transformation models} (ATMs; Section~\ref{sec:ats}) that learn expressive distributions based on interpretable parametric transformations. ATMs can be seen as a generalization of autoregressive models. We study the autoregressive transformation of order $p$ (AT($p$)) in Section~\ref{sec:atp} as the closest neighbor to a parametric autoregressive model, and derive asymptotic results for estimated parameters in Section~\ref{sec:mle}. Finally, we provide evidence for the efficacy of our proposal both with numerical experiments based on simulated data and by comparing ATMs against other existing time series methods. 

\section{Background and Related Work} \label{sec:background}

Approaches that model the conditional density can be distinguished by their underlying distribution assumption. Approaches can be parametric, such as mixture density networks \citep{Bishop.1994} for conditional density estimation and then learn the parameters of a pre-specified parametric distribution or non-parametric such as Bayesian non-parametrics \citep{Dunson.2010}. A third line of research that we describe as semi-parametric, are approaches that start with a simple parametric distribution assumption $F_Z$ and end up with a far more flexible distribution $F_{Y\mid \bm{x}}$ by transforming $F_Z$ (multiple times). Such approaches have sparked great interest in recent years, triggered by research ideas such as density estimation using non-linear independent components estimation or real-valued non-volume preserving transformations \citep{Dinh.2017}. A general notion of such transformations is known as normalizing flow \citep[NF;][]{Papamakarios.2019}, where realizations $\bm{z} \sim F_Z$ of an error distribution $F_z$ are transformed to  observations $\bm{y}$ via
\begin{align}
\bm{y} &= h_k \circ h_{k-1} \circ \cdots \circ h_1(\bm{z})
\end{align}
using $k$ transformation functions. Many different approaches exist to define expressive flows. These are often defined as a chain of several transformations or an expressive neural network and allow for universal representation of $F_{Y\mid \bm{x}}$ \citep{Papamakarios.2019}. Autoregressive models \citep[e.g.,][]{Bengio.1999, Uria.2016} for distribution estimation of continuous variables are a special case of NFs, more precisely autoregressive flows \citep[AFs;][]{Kingma.2016, Papamakarios.2017}, with a single transformation. 

\paragraph{Transformation models} Transformation models \citep[TMs;][]{Hothorn.2014}, a similar concept to NFs, only consist of a single transformation and thereby better allow theoretically studying model properties. The transformation in TMs is chosen to be expressive enough on its own and comes with desirable approximation guarantees. Instead of a transformation from $\bm{z}$ to $\bm{y}$, TMs define an inverse flow $h(\bm{y}) = \bm{z}$. The key idea of TMs is that many well-known statistical regression models can be represented by a base distribution $F_Z$ and some transformation function $h$. Prominent examples include linear regression or the Cox proportional hazards model \citep{Cox.1972}, which can both be seen as a special case of TMs \citep{Hothorn.2014}. Various authors have noted the connection between autoregressive models and NFs \citep[e.g.,][]{Papamakarios.2019} and between TMs and NFs \citep[e.g.,][]{Sick.2020}. Advantages of TMs and conditional TMs (CTMs) are their parsimony in terms of parameters, interpretability of the input-output relationship, and existing theoretical results \citep{mlt.2018}. While mostly discussed in the statistical literature, various recent TM advancements have been also proposed in the field of machine learning \citep[see, e.g.,][]{Van.2011} and deep learning \citep[see, e.g.,][]{Baumann.2020, Kook.2020, Kook.2022}. 


\paragraph{Time series forecasting}

In time series forecasting, many approaches rely on autoregressive models, with one of the most commonly known linear models being autoregressive (integrated) moving average (AR(I)MA) models \citep[see, e.g.,][]{Shumway.2020}. Extensions include the bilinear model of \cite{granger1978invertibility,rao1981theory}, or the Markov switching autoregressive model by \cite{hamilton2010regime}. Related to these autoregressive models are stochastic volatility models \citep{kastner2017} building upon the theory of stochastic processes. In probabilistic forecasting, Bayesian model averaging \citep{raftery2005using} and distributional regression forecasting \citep{Schlosser_2019} are two further popular approaches while many other Bayesian and non-Bayesian techniques exist \citep[see, e.g.,][for an overview]{gneiting2014probabilistic}.

\subsection{Transformation models}

Parametrized transformation models as proposed by \citet{Hothorn.2014, mlt.2018} are likelihood-based approaches to estimate the CDF $F_{Y}$ of $Y$. The main ingredient of TMs is a monotonic transformation function $h$ to convert a simple base distribution $F_Z$ to a more complex and appropriate CDF $F_{Y}$. Conditional TMs (CTMs) work analogously for the conditional distribution of $Y$ given features $\bm{x}\in\chi$ from feature space $\chi$:
\begin{equation} \label{eq:CTM}
F_{Y\mid \bm{x}}(y) = \mathbb{P}(Y \leq y \mid  \bm{x}) = F_Z(h(y \mid  \bm{x})).    
\end{equation}
CTMs learn $h(y \mid  \bm{x})$ from the data, i.e., estimate a model for the (conditional) aleatoric uncertainty. 
A convenient parameterization of $h$ for continuous $Y$ are Bernstein polynomials \citep[BSPs;][]{Farouki.2012} with order $M$ (usually $M \ll 50$). BSPs are motived by the Bernstein approximation \citep{Bernstein.1912} with uniform convergence guarantees for $M\to\infty$, while also being computationally attractive with only $M+1$ parameters. BSPs further have easy and analytically accessible derivatives, which makes them a particularly interesting choice for the change of random variables. We denote the BSP basis by $\bm{a}_M: \Xi \mapsto \mathbb{R}^{M+1}$ with sample space $\Xi$. The transformation $h$ is then defined as $h(y\mid \bm{x}) = \bm{a}_M(y)^\top \bm{\vartheta}(\bm{x})$ with feature-dependent basis coefficients $\bm{\vartheta}$. This can be seen as an evaluation of $y$ based on a mixture of Beta densities $f_{Be(\kappa,\mu)}$ with different distribution parameters $\kappa,\mu$ and weights $\bm{\vartheta}(\bm{x})$:
\begin{equation} \label{eq:BSP}
 \bm{a}_M(y)^\top \bm{\vartheta}(\bm{x}) = \frac{ \sum_{m = 0}^{M} \vartheta_m(\bm{x}) f_{Be(m + 1, M - m + 1)}(\tilde{y})}{M+1},
\end{equation}
where $\tilde{y}$ is a rescaled version of $y$ to ensure $\tilde{y} \in [0,1]$. Restricting $\vartheta_m > \vartheta_{m-1}$ for $m = 1, \ldots,  M + 1$ guarantees monotonicity of $h$ and thus of the estimated CDF. Roughly speaking, using BSPs of order $M$, allows to model the polynomials of degree $M$ of $y$.

\subsection{Model definition} \label{sec:modelint}

The transformation function $h$ can include different data dependencies. One common choice \citep{Hothorn.2020,Baumann.2020} is to split the transformation function into two parts 
\begin{equation} \label{eq:TMs}
h(y \mid  \bm{x}) = h_1(y,\bm{x}) + h_2(\bm{x}) = \bm{a}(y)^\top \bm{\vartheta}(\bm{x}) + \beta(\bm{x}),    
\end{equation}
where $\bm{a}(y)$ is a pre-defined basis function such as the BSP basis (omitting $M$ for readability in the following), $\bm{\vartheta} : \chi_\vartheta \mapsto \mathbb{R}^{M+1}$ a conditional parameter function defined on $\chi_\vartheta \subseteq \chi$ and $\beta(\bm{x})$ models a feature-induced shift in the transformation function. The flexibility and interpretability of TMs stems from the parameterization 
\begin{equation} \label{eq:vartheta}
 \bm{\vartheta}(\bm{x}) =  \sum_{j=1}^J \bm{\Gamma}_j. \bm{b}_j(\bm{x}),  
\end{equation}
where the matrix $\bm{\Gamma}_j \in \mathbb{R}^{(M+1) \times O_j}, O_j \geq 1,$ subsumes all trainable parameters and represents the effect of the interaction between the basis functions in $\bm{a}$ and the chosen predictor terms $\bm{b}_j : \chi_{b_j} \mapsto \mathbb{R}^{O_j}, \chi_{b_j} \subseteq \chi$. The predictor terms $\bm{b}_j$ have a role similar to base learners in boosting and represent simple learnable functions. For example, a predictor term can be the $j$th feature, $\bm{b}_j(\bm{x}) = x_{j}$, and $\bm{\Gamma}_j \in \mathbb{R}^{(M+1) \times 1}$ describes the linear effect of this feature on the $M+1$ basis coefficients, i.e., how the feature $x_{j}$ relates to the density transformation from  $Z$ to $Y\mid \bm{x}$. Other structured non-linear terms such as splines allow for interpretable lower-dimensional non-linear relationships. Various authors also proposed neural network (unstructured) predictors to allow potentially multidimensional feature effects or to incorporate unstructured data sources \citep{Sick.2020, Baumann.2020, Kook.2020}. In a similar fashion, $\beta(\bm{x})$ 
can be defined using various structured and unstructured predictors. 

\paragraph{Interpretability} Relating features and their effect in an additive fashion allows to directly assess the impact of each feature on the transformation and also whether changes in the feature just shift the distribution in its location or if the relationship also transforms other distribution characteristics such as variability or skewness \citep[see, e.g.,][for more details]{Baumann.2020}. 

\paragraph{Relationship with autoregressive flows} 

In the notation of AFs, $h^{-1}(\cdot)$ is known as \emph{transformer}, a parameterized and bijective function. By the definition of \eqref{eq:TMs}, the transformer in the case of TMs is represented by the basis function $\bm{a}(\cdot)$ and parameters $\bm{\vartheta}$. In AFs, these transformer parameters are learned by a \emph{conditioner}, which in the case of TMs are the functions $\bm{b}_j$. In line with the assumptions made for AFs, these conditioners in TMs do not need to be bijective functions themselves. 

\section{Autoregressive Transformations} \label{sec:ats}

Inspired by TMs and AFs, we propose autoregressive transformation models (ATMs). Our work is the first to adapt TMs for time series data and thereby lays the foundation for future extensions of TMs for time series forecasting. The basic idea is to use a parameter-free base distribution $F_Z$ and transform this distribution in an interpretable fashion to obtain $F_{Y\mid \bm{x}}$. One of the assumptions of TMs is the stochastic independence of observations, i.e., $Y_i\mid \bm{x}_i \bot Y_j\mid \bm{x}_j, i\neq j$. When $Y$ is a time series, this assumption does clearly not hold. In contrast, this assumption is not required for AFs. 

Let $t \in \mathcal{T} \subseteq \mathbb{N}_0$ be a time index for the time series $(Y_t)_{t\in\mathcal{T}}$. Assume 
\begin{equation}
Y_t \mid  \mathcal{F}_{t-1} \sim G(Y_{t-1},\ldots,Y_{t-p}; \bm{\theta})    
\end{equation}
for some $p \in \{1,\ldots,t\}$, distribution $G$, parameter $\bm{\theta} \in \Theta$ with compact parameter space $\Theta \subset \mathbb{R}^v$ and filtration $\mathcal{F}_s$, $s\in \mathcal{T}$, $s < t$, on the underlying probability space.  Assume that the joint distribution of $Y_t, Y_{t-1}, \ldots, Y_1$ possesses the Markov property with order $p$, i.e., the joint distribution, expressed through its absolutely continuous density $f$, can be rewritten as product of its conditionals with $p$ lags:
\begin{equation} \label{eq:markov}
f(y_t, \ldots,y_1\mid \bm{x}) = \prod_{s=p+1}^t f(y_s\mid y_{s-1},\ldots,y_{s-p}, \bm{x}).    
\end{equation}
We use $\bm{x}$ to denote (potentially time-varying) features that are additional (exogenous) features. Their time-dependency is omitted for better readability here and in the following. Given this autoregressive structure, we propose a time-dependent transformation $h_t$ that extends (C)TMs to account for filtration and time-varying feature information. By modeling the conditional distribution of all time points in a flexible manner, ATMs provide an expressive way to account for aleatoric uncertainty in the data.
\begin{definition}{\textbf{Autoregressive Transformation Models}}
Let $h_t, t\in\mathcal{T}$, be a time-dependent monotonic transformation function and $F_Z$ the parameter-free base distribution as in Definition~1 in the Supplementary Material. We define autoregressive transformation models as follows:
\begin{equation} \label{eq:ATs}
\begin{split}
    \mathbb{P}(Y_t \leq y_t \mid  \mathcal{F}_{t-1}, \bm{x}) &= F_{Y_t\mid  \mathcal{F}_{t-1}, \bm{x}}(y_t)\\ &= F_Z(h_t(y_t\mid  \mathcal{F}_{t-1}, \bm{x})).
\end{split}
\end{equation}
\end{definition}
This can be seen as the natural extension of \eqref{eq:CTM} for time series data with autoregressive property and time-varying transformation function $h_t$. In other words, \eqref{eq:ATs} says that after transforming $y_t$ with $h_t$, its conditional distribution follows the base distribution $F_Z$, or vice versa, a random variable $Z \sim F_Z$ can be transformed to follow the distribution $Y_t\mid \bm{x}$ using $h_t^{-1}$.

\paragraph{Relationship with autoregressive models and autoregressive flows} Autoregressive models \citep[AMs;][]{Bengio.1999} and AFs both rely on the factorization of the joint distribution into conditionals as in \eqref{eq:markov}. Using the CDF of each conditional in \eqref{eq:markov} as transformer in an AF, we obtain the class of AMs \citep{Papamakarios.2019}. AMs and ATMs are thus both (inverse) flows using a single transformation, but with different transformers and, as we will outline in Section~\ref{sec:subclass}, also with different conditioners.

\subsection{Likelihood-based estimation}

Based on \eqref{eq:markov},~\eqref{eq:ATs} and the change of variable theorem, the likelihood contribution of the $t$th observation $y_t$ in ATMs is given by
\begin{equation*} 
\begin{split}
&f_{Y\mid x}(y_t\mid  \mathcal{F}_{t-1},\bm{x}) =\\
& f_Z(h_t(y_t\mid \mathcal{F}_{t-1}, \bm{x})) \cdot \bigl\lvert \frac{\partial h_t(y_t\mid \mathcal{F}_{t-1},\bm{x})}{\partial y_t} \bigr\rvert
\end{split}
\end{equation*}
and the full likelihood for $T$ observations thus by
\begin{equation} \label{eq:like}
\begin{split}
&f_{Y\mid x}(Y_T,\ldots,Y_1\mid \mathcal{Y}_0,\bm{x}) =\\ 
&\prod_{t=1}^T \left\{ f_Z(h_t(y_t\mid \mathcal{F}_{t-1}, \bm{x})) \cdot \bigl\lvert \frac{\partial h_t(y_t\mid \mathcal{F}_{t-1},\bm{x})}{\partial y_t} \bigr\rvert \right\},
\end{split}
\end{equation}
where $\mathcal{Y}_0 = (y_0, \ldots, y_{-p+1})$ are known finite starting values and $\mathcal{F}_0$ only contains these values. Based on \eqref{eq:like}, we define the loss of all model parameters $\bm{\theta}$ as negative log-likelihood $-\ell(\bm{\theta}) := - \log f_{Y\mid x}(Y_T,\ldots,Y_1\mid \mathcal{Y}_0,\bm{x})$ given by
\begin{equation} \label{eq:loglike}
\begin{split}
    - \sum_{t=1}^T \Bigl\{ \log  f_Z(h_t(y_t\mid \mathcal{F}_{t-1}, \bm{x})) +  \\ 
 \log \bigl\vert \frac{\partial h_t(y_t\mid \mathcal{F}_{t-1},\bm{x})}{\partial y_t} \bigr\vert \Bigr\},
\end{split}
\end{equation}
and use \eqref{eq:loglike} to train the model. 

As for AFs, many special cases can be defined from the above definition and more concrete structural assumptions for $h_t$ make ATMs an interesting alternative to other methods in practice. We will elaborate on meaningful structural assumptions in the following. 



\subsection{Structural assumptions} \label{sec:subclass}

\begin{figure}
    \centering
    \includegraphics[width=0.48\textwidth]{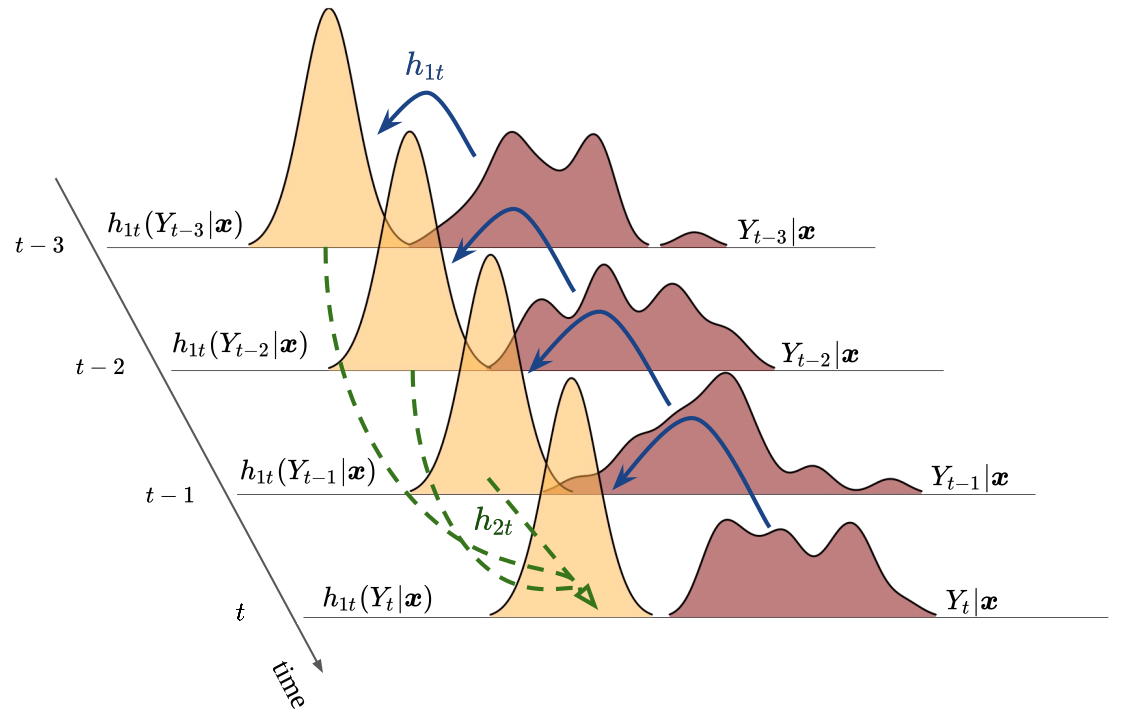}
    \caption{Illustration of a transformation process induced by the structural assumption of Section~\ref{sec:subclass}. The original data history $\mathcal{F}_{t-1}$ (red) is transformed into a base distribution (orange) using the transformation $h_{1t}$ (solid blue arrow) and then further transformed using $h_{2t}$ (dashed green arrow) to match the transformed distribution of the current time point $t$.}
    \label{fig:intuition}
\end{figure}

In CTMs, the transformation function $h$ is usually decomposed as $h(y \mid  \bm{x}) = h_1(y \mid  \bm{x}) + h_2(\bm{x})$, where $h_1$ is a function depending on $y$ and $h_2$ is a transformation-shift function depending only on $\bm{x}$. 
For time-varying transformations $h_t$ our fundamental idea is that the outcome $y_t$ shares the same transformation with its filtration $\mathcal{F}_{t-1}$, i.e., the lags $\mathcal{Y}_t = (y_{t-1}, \ldots, y_{t-p})$. In other words, a transformation applied to the outcome must be equally applied to its predecessor in time to make sense of the autoregressive structural assumption. An appropriate transformation structure can thus be described by 
\begin{equation} \label{eq:sats}
\begin{split}
    h_t&(y_t \mid  \mathcal{F}_{t-1}, \bm{x})\\ &= h_{1t}(y_t \mid  \bm{x}) + h_{2t}((h_{1t} \odot \mathcal{Y}_{t}\mid  \mathcal{F}_{t-1}, \bm{x}) \mid  \bm{x})\\ &=: \lambda_{1t} + \lambda_{2t},
\end{split}
\end{equation}
for $t\in \mathcal{T}$, where $\odot$ indicates the element-wise application of $h_{1t}$ to all lags in $\mathcal{Y}_t$. In other words, ATMs first apply the same transformation $h_{1t}$ to $y_t$ and individually to $y_{t-1},y_{t-2},\ldots$, and then further consider a transformation function $h_{2t}$ to shift the distribution (and thereby potentially other distribution characteristics) based on the transformed filtration. While the additivity assumption of $\lambda_{1t}$ and $\lambda_{2t}$ seems restrictive at first glance, the imposed relationship between $y_t$ and $\mathcal{Y}_{t}$ only needs to hold in the transformed probability space. For example, $h_{1t}$ can compensate for a multiplicative autoregressive effect between the filtration and $y_t$ by implicitly learning a $\log$-transformation (cf. Section~\ref{sec:exp_arp}). At the same time, the additivity assumption offers a nice interpretation of the model, also depicted in Figure~\ref{fig:intuition}: After transforming $y_t$ and $\mathcal{Y}_{t}$, \eqref{eq:sats} implies that training an ATM is equal to a regression model of the form $\lambda_{1t} = \lambda_{2t} + \varepsilon$, with additive error term $\varepsilon \sim F_Z$ (cf. Proposition~1 in Supplementary Material~A.2). This also helps explaining why only $\lambda_{2t}$ depends on $\mathcal{F}_{t-1}$: if $\lambda_{1t}$ also involves $\mathcal{F}_{t-1}$, ATMs would effectively model the joint distribution of the current time point and the whole filtration, which in turn contradicts the Markov assumption \eqref{eq:markov}.

Specifying $h_{1t}$ very flexible clearly results in overfitting. As for CTMs, we use a feature-driven basis function representation $h_{1t}(y_t  \mid  \bm{x}) = \bm{a}(y_t)^\top \bm{\vartheta}(\bm{x})$ with BSPs $\bm{a}$ and 
specify their weights as in \eqref{eq:vartheta}. 
The additional transformation $h_{2t}$ ensures enough flexibility for the relationship between the transformed response and the transformed filtration, e.g., by using a non-linear model or neural network. An interesting special case arises for linear transformations in $h_{2t}$, which we elaborate in Section~\ref{sec:atp} in more detail. 

\paragraph{Interpretability} The three main properties that make ATMs interpretable are 1) their additive predictor structure as outlined in \eqref{eq:vartheta}; 2) the clear relationship between features and the outcome through the BSP basis, and 3) ATM's structural assumption as given in \eqref{eq:sats}. As for (generalized) linear models, the additivity assumption in the predictor allows interpreting feature influences through their partial effect ceteris paribus. On the other hand, choices of $M$ and $F_Z$ will influence the relationship between features and outcome by inducing different types of models. A normal distribution assumption for $F_Z$ and $M = 1$ will turn ATMs into an additive regression model with Gaussian error distribution (see also Section~\ref{sec:atp}). For $M > 1$, features in $h_1$ will also influence higher moments of $Y\mid \bm{x}$ and allow more flexibility in modeling $F_{Y\mid \bm{x}}$. For example, a (smooth) monotonously increasing feature effect will induce rising moments of $Y\mid \bm{x}$ with increasing feature values.  Other choices for $F_Z$ such as the logistic distribution also allow for easy interpretation of feature effects \citep[e.g., on the log-odds ratio scale; see][]{Kook.2020}. Finally, the structural assumption of ATMs enforces that the two previous interpretability aspects are consistent over time. We will provide an additional illustrative example in Section~\ref{sec:bench}, further explanations in Supplementary Material~B, and refer to \citet{Hothorn.2014} for more details on interpretability of CTMs.

\paragraph{Implementation} In order to allow for a flexible choice of transformation functions and predictors $\bm{b}_j$, we propose to implement ATMs in a neural network and use stochastic gradient descent for optimization. While this allows for complex model definitions, there are also several computational advantages. In a network, weight sharing for $h_{1t}$ across time points is straightforward to implement and common optimization routines such as Adam \citep{Kingma.2014} prove to work well for ATMs despite the monotonicity constraints required for the BSP basis. Furthermore, as basis evaluations for a large number of outcome lags in $\mathcal{F}_{t-1}$ can be computationally expensive for large $p$ (with space complexity $\mathcal{O}(t \cdot M \cdot p)$) and add $M$ additional columns per lag to the feature matrix, an additional advantage is the dynamic nature of mini-batch training. In this specific case, it allows for evaluating the bases only during training and separately in each mini-batch. It is therefore never required to set up and store the respective matrices. 



\section{AT($p$) Model} \label{sec:atp}

A particular interesting special case of ATMs is the AT($p$) model. This model class is a direct extension of the well-known autoregressive model of order $p$ \citep[short AR($p$) model;][]{Shumway.2020} to transformation models. 
\begin{definition}{\textbf{AT($p$)~model}}
We define the AT($p$) model, a special class of ATMs, by setting $h_{1t}(y_t\mid \bm{x}) = \bm{a}(y_t)^\top \bm{\vartheta}(\bm{x})$, and $h_{2t}(\mathcal{F}_{t-1},\bm{x}) = \sum_{j=1}^p \phi_j h_{1t}(y_{t-j}) + r(\bm{x})$, i.e., an autoregressive shift term with optional exogenous remainder term $r(\bm{x})$.
\end{definition}
As for classical time series approaches, $\phi_j$ are the regression coefficients relating the different lags to the outcome and $r$ is a structured model component (e.g., linear effects) of exogenous features that do not vary over time.

\subsection{Model Details}

The AT($p$) model is a very powerful and interesting model class for itself, as it allows to recover the classical time series AR($p$) model when setting $M=1$, $\bm{\vartheta}(\bm{x}) \equiv \bm{\vartheta}$ and $r(\bm{x}) \equiv 0$ (see Proposition~2 in Supplementary Material~A for a proof of equivalence). But it can also be extended to more flexible autoregressive models in various directions. We can increase $M$ to get a more flexible density, allowing us to deviate from the base distribution assumption $F_Z$, e.g., to relax the normal distribution assumption of AR models. Alternatively, incorporating exogenous effects into $h_{1t}$ allows to estimate the density data-driven or to introduce exogenous shifts in time series using features $\bm{x}$ in $r(\bm{x})$. ATMs can also recover well-known transformed autoregressive models such as the multiplicative autoregressive model \citep{Wong.2000} as demonstrated in Section~\ref{sec:exp_arp}. When specifying $M$ large enough, an AT($p$) model will, e.g., learn the log-transformation function required to transform a multiplicative autoregressive time series to an additive autoregressive time series on the log-scale. In general, this allows the user to learn autoregressive models without the need to find an appropriate transformation before applying the time series model. This means that the uncertainty about preprocessing steps \citep[e.g., a Box-Cox transformation;][]{Sakia.1992} is incorporated into the model estimation, making parts of the pre-processing obsolete for the modeler and its uncertainty automatically available. 


Non-linear extensions of AT($p$) models can be constructed by modeling $\mathcal{Y}_t$ in $h_{2t}$ non-linearly, allowing ATMs to resemble model classes such as non-linear AR models with exogenous terms \citep[e.g.,][]{Lin.1996}. In practice, values for $p$ can, e.g., be found using a (forward) hyperparameter search by comparing the different model likelihoods.

\subsection{Asymptotic theory} \label{sec:theory} \label{sec:mle}

\begin{figure}[!t]
    \includegraphics[width = 0.9\columnwidth]{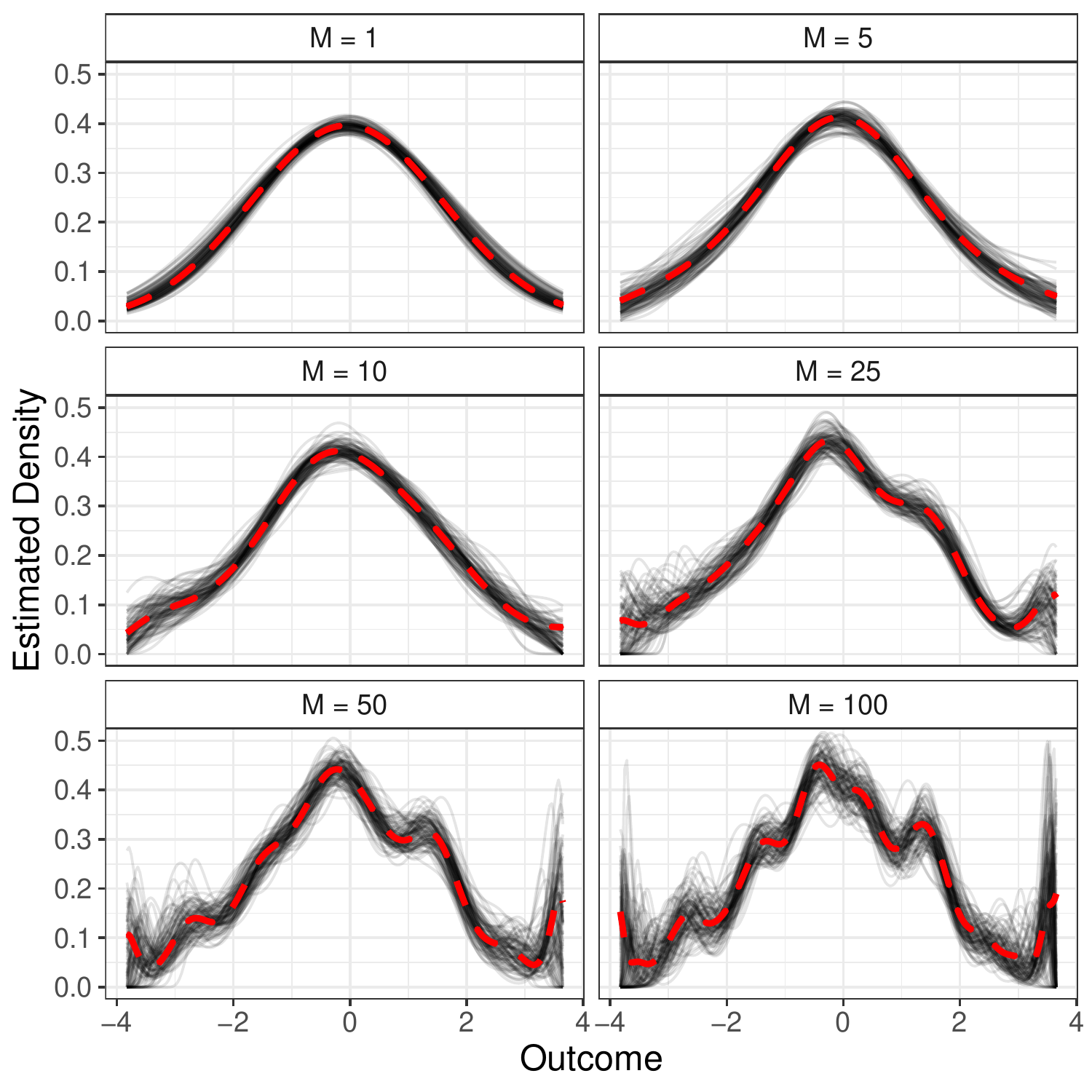}
    \caption{Aleatoric vs. epistemic uncertainty: Different plots correspond to different orders of the BSP basis $M$, inducing different amounts of expressiveness and aleatoric uncertainty. In each plot, the fitted density is shown in red, and model uncertainties of this density based on the epistemic uncertainty in black. Epistemic uncertainty is generated according to results in Theorem~\ref{theorem:norm} and~\ref{theorem:est}.}
    \label{fig:aleaepi}
\end{figure}

An important yet often neglected aspect of probabilistic forecasts is the epistemic uncertainty, i.e., the uncertainty in model parameters. Based on general asymptotic theory for time series models \citep{Ling.2010}, we derive theoretical properties for AT($p$)s in this section. 

Let $\bm{\theta}^\ast$ be the true value of $\bm{\theta}$ and interior point of $\Theta$. We define the following quantities involved in standard asymptotic MLE theory: Let $\hat{\bm{\theta}}_T = \arg\min_\Theta -\ell(\bm{\theta})$ be the parameter estimator based on Maximum-Likelihood estimation (MLE), $\nabla_T(\bm{\theta}) = \partial \ell_T(\bm{\theta})/\partial \bm{\theta}$, $\mathcal{J}_T(\bm{\theta}) = -\partial^2 \ell_T(\bm{\theta}) /(\partial\bm{\theta} \partial \bm{\theta}^\prime)$, $\mathcal{I} = \mathbb{E}_G(\mathcal{J}_T(\bm{\theta}^\ast))$ and $\mathfrak{J} = \mathbb{E}_G(\nabla_T(\bm{\theta}^\ast) \nabla^\top_T(\bm{\theta}^\ast))$. We further state necessary assumptions to apply the theory of \citet{Ling.2010} for a time series $(Y_t)_{t\in\mathcal{T}}$ with known initial values $\mathcal{Y}_0$ as defined in Section~\ref{sec:ats}.

\begin{assumption}
\label{assum:th1} Assume 
\begin{itemize}
\item[(i)] $(Y_t)_{t\in\mathcal{T}}$ is strictly stationary and ergodic;
    \item[(ii)] $\mathbb{E}_G \{ \sup_{\bm{\theta} \in \Theta} [\ell_T(\bm{\theta})] \} < \infty$ and $\bm{\theta}^\ast$ is unique;
    \item[(iii)] $\nabla_T(\bm{\theta}^\ast)$ is a martingale difference w.r.t. $\mathcal{F}_{T-1}$ with $0 < \mathfrak{J} < \infty$;
    \item[(iv)] $\mathcal{I}$ is positive-definite and for some $\xi > 0$ $\mathbb{E}_G \{ \sup_{\bm{\theta}: ||\bm{\theta}-\bm{\theta}^\ast||<\xi} || \mathcal{J}_T(\bm{\theta} || \} < \infty$.
\end{itemize}
\end{assumption}

Assumptions~\ref{assum:th1} are common assumptions required for many time series models. We require only these and no other assumptions since AT($p$)s and non-linear extensions are fully-parameterized time series models. This allows us to derive general statistical inference theory for AT($p$) models.

\begin{theorem}[Consistency]
\label{theorem:cons}
If elements in $\mathcal{Y}_0$ are finite and Assumption~\ref{assum:th1}(i) holds, then $\hat{\bm{\theta}}_T \overset{ a.s.}{\longrightarrow} \bm{\theta}^\ast$ for $T \to \infty$. 
\end{theorem}

As stated in \citet{mlt.2018}, Assumption~\ref{assum:th1}(ii) holds if $\bm{a}$ is not arbitrarily ill-posed. In practice, both a finite $\mathcal{Y}_0$ and Assumption~\ref{assum:th1}(i) are realistic assumptions. Making two additional and also rather weak assumptions (1(iii)-(iv)) allows to derive the asymptotic normal distribution for $\hat{\bm{\theta}}$.

\begin{theorem}[Asymptotic Normality]
\label{theorem:norm}
If $y_0$ is finite and Assumptions~\ref{assum:th1} hold, then  for $T \to \infty$,
$$
\hat{\bm{\theta}}_T = \bm{\theta}^\ast + {O}(\sqrt{(\log\log T) / T})$$ 
and 
$$\sqrt{T}(\hat{\bm{\theta}}_T - \bm{\theta}^\ast) \overset{D}{\longrightarrow} \mathcal{N}(0, \mathcal{I}^{-1} \mathfrak{J} \mathcal{I}^{-1}).
$$
\end{theorem}

Based on the same assumptions, a consistent estimator for the covariance can be derived.

\begin{theorem}[Consistent Covariance Estimator]
\label{theorem:est}
For finite $y_0$ and under Assumptions~\ref{assum:th1},  
$$\hat{\mathcal{I}}_T = \frac{1}{T} \sum_{t=1}^T \mathcal{J}_T(\hat{\bm{\theta}}_T) \,\, \text{and} \,\,\, \hat{\mathfrak{J}}_T = \frac{1}{T} \sum_{t=1}^T \nabla_T(\hat{\bm{\theta}}_T) \nabla^\top_T(\hat{\bm{\theta}}_T)$$
are consistent estimators for $\mathcal{I}$ and $\mathfrak{J}$, respectively.
\end{theorem}

The previous theorems can be proven by observing that the AT($p$) model structure and all made assumptions follow the general asymptotic theory for time series models as given in \citet{Ling.2010}. See Supplementary Material~A for details.

Using the above results, we can derive statistically valid UQ. An example is depicted in Figure~\ref{fig:aleaepi}. 
Since $h$ is parameterized through $\bm{\theta}$, it is also possible to derive the so-called structural uncertainty of ATMs, i.e., the uncertainty induced by the discrepancy between the model's CDF $F_{Y\mid x}(y\mid \bm{x};\bm{\theta})$ and the true CDF $F^\ast_{Y\mid x}(y\mid \bm{x})$ \citep{Liu.2019}. More specifically, $h$ can be represented using a linear transformation of $\bm{\theta}$, $h = \bm{\Upsilon} \bm{\theta}$, implying the (co-)variance $\bm{\Upsilon} \mathcal{I}^{-1} \mathfrak{J}(\bm{\theta}^\ast) \mathcal{I}^{-1} \bm{\Upsilon}^\top$ for $\hat{h}$. 

\paragraph{Practical application} ATM define the distribution $F_{Y_t\mid \mathcal{F}_{t-1},x}$ via $F_{Y\mid \mathcal{F}_{t-1},x} = F_Z \circ h_t$, where $h_t$ is parameterized by $\bm{\theta}$. In order to assess parameter uncertainty in the estimated density as, e.g. visualized in Figure~\ref{fig:motivating} and~\ref{fig:aleaepi}, we propose to use a parametric Bootstrap described in detail in Supplementary Material~C.

\section{Experiments} \label{sec:numexp}

\begin{figure}[ht]
    \centering
    \includegraphics[width=0.45\textwidth]{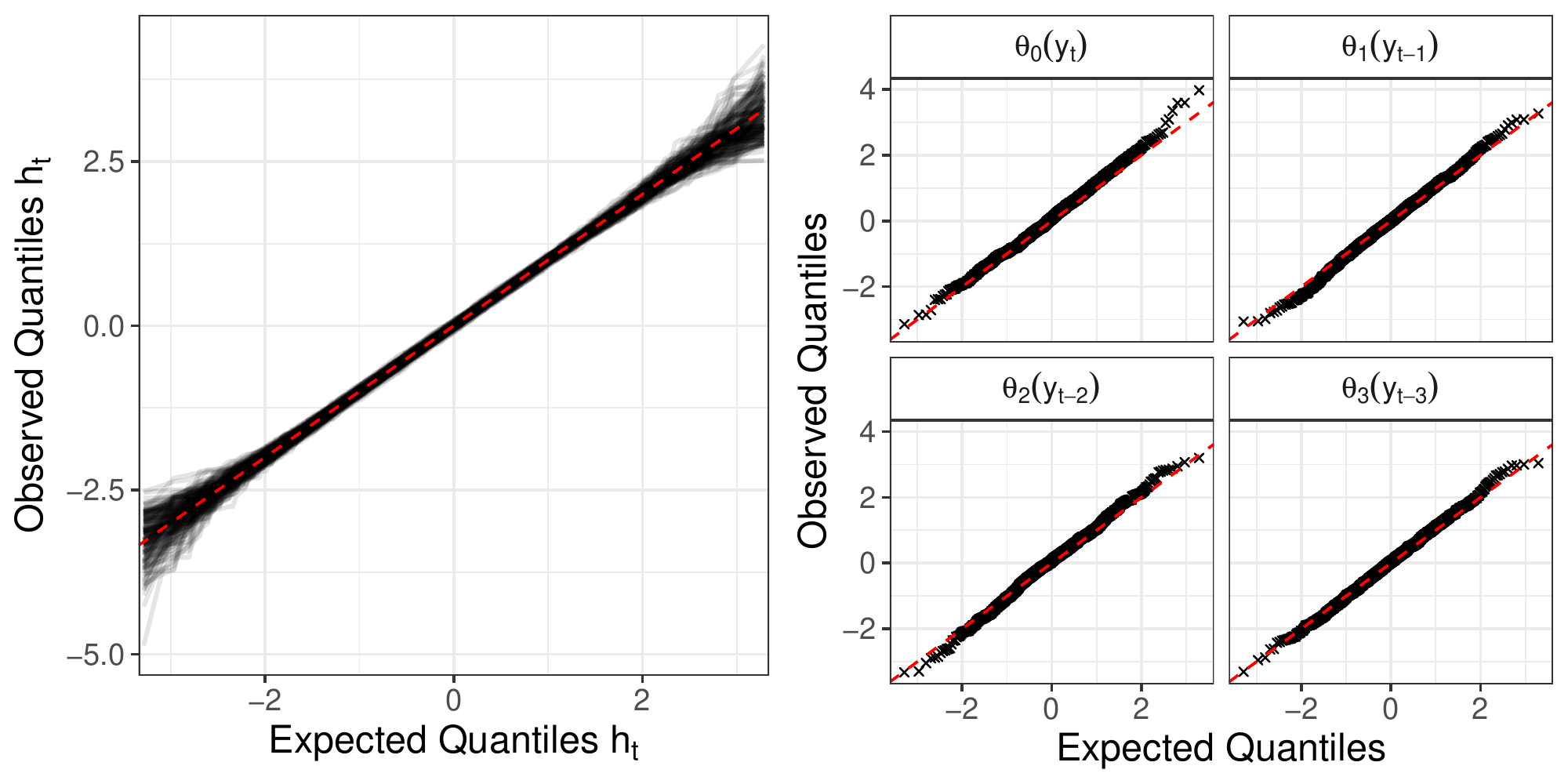}
    \caption{Empirical evidence for the correctness of our theoretical results on PU: Expected vs.~observed quantiles of the transformation function $h_t$ (left; one line per dataset) and model parameters $\bm{\theta}$ for the different (lagged) transformed outcomes (right; one cross per dataset) based on 1000 simulation replications. The ideal angle bisector is plotted in red.}
    \label{fig:qq}
\end{figure}

We will first investigate theoretical properties of ATMs and the validity of statistical inference statements using simulation studies. We then compare our approach against other state-of-the-art methods described in the previous section on probabilistic forecasting tasks in a benchmark study. Additional results can be found in the Supplementary Material~D.

\subsection{Simulation Study} \label{sec:exp_arp}

\setlength{\tabcolsep}{4pt}
\begin{table}[ht]
\centering
\small
\caption{Average and standard deviation (brackets) of the MSE (multiplied by $100$ for better readability) between estimated and true coefficients in an AR($p$) model using our approach on the tampered data (bottom row) and the corresponding oracle based on the true data (Oracle).}  \label{tab:sats}
\begin{tabular}{ccccc}
 & T & $p=1$ & $p=2$ & $p=4$ \\ 
  \hline
Oracle & \multirow{ 2}{*}{$400$} &    0.33 (0.31) & 0.22 (0.19) & 0.25 (0.13) \\ 
AT($p$) &  & 0.52 (0.46) & 0.33 (0.3) & 0.34 (0.23) \\ 
 Oracle  & \multirow{ 2}{*}{$800$} &   0.27 (0.34) & 0.13 (0.12) & 0.13 (0.085) \\ 
AT($p$) &  & 0.26 (0.36) & 0.17 (0.17) & 0.18 (0.12)  \\ 
   \hline
\end{tabular}
\end{table}

\paragraph{Equivalence and consistency} We first demonstrate Theorem~\ref{theorem:cons} and Proposition~2 in the Supplementary Material, i.e., for growing number of observations AT($p$) models can recover AR($p$) models when equally specified. We therefore simulate various AR models using lags $p\in \{1,2,4\}$, $T\in\{200,400,800\}$ and estimate both a classical AR($p$) model and an AT($p$) model for 20 replications. For the latter, we use the mapping derived in Proposition~2 to obtain the estimated AR coefficients from the AT($p$) model. In Table~D1 in the Supplementary Material~D we compare both models based on their estimated coefficients against the ground truth using the mean squared error (MSE). Results show that the AT($p$) model can empirically recover the AR($p$) model very well. 

\paragraph{Flexibility} Next, we demonstrate how the AT($p$) model with $M = 30$ can recover a multiplicative autoregressive process. We therefore generate data using an AR model with different lags $p$ and observations $n$ as before. This time, however, we provide the AT($p$) model only with the exponentiated data $\check{y}_t = \exp(y_t)$. This means the model needs to learn the inverse transformation back to $y_t$ itself. Despite having to estimate the log-transformation in addition, the AT($p$) model recovers the true model well and, for larger $n$, is even competitive to the ground truth model (Oracle) that has access to the original non-exponentiated data (cf. Table~D2 for an excerpt of the results).  

\paragraph{Epistemic Uncertainty} In this experiment we validate our theoretical results proposed in Section~\ref{sec:mle}. As in the previous experiment, we try to learn the log-transformed AR model using an AT($p=3$) model with coefficients $(0.3, 0.2, 0.1)$. After estimation, we check the empirical distribution of $\hat{\bm{\theta}}$ and $\hat{h}$ against their respective theoretical one in 1000 simulation replications. Figure~\ref{fig:qq} depicts a quantile-quantile plot of the empirical and theoretical distribution for both $h$ and all 4 parameters (intercept and three lag coefficients). The empirical distributions are well aligned with their theoretical distribution as derived in Section~\ref{sec:mle}, confirming our theoretical results.



\subsection{Benchmarks} \label{sec:bench}
Finally, we compare our approach to its closest neighbor in the class of additive models, the  ARIMA model \citep{forecast}, against a simple Box-Cox transformation (BoxCox), a neural network for mean-variance estimation (MVN) and a mixture density network \citep[MDN;][]{Bishop.1994}.  While there are many further forecasting techniques, especially in deep learning, we purposely exclude more complex machine and deep learning approaches to compare AT($p$)s with approaches of similar complexity. More specifically, the different competitors were chosen to derive the following insights: The comparison of the AT($p$) model with the ARIMA model will indicate whether relaxing the parametric assumption using TMs can improve performance while both methods take time series lags into account. The comparison of our method with BoxCox, on the other hand, will show similar performance if there is no relevant information in the lags of the time series. The MVN can potentially learn time series-specific variances but is not given the lagged information as input. A good performance of the MVN will thus indicate heteroscedasticity in the data generating process which can, however, be accounted for using a parametric distributional regression approach. Finally, the MDN is an alternative approach to the AT($p$) model that tries to overcome the parametric assumption by modeling a mixture of normal distributions.

\paragraph{Hyperparameter Setup} We define the AT($p$) model by using an unconditional $\bm{\vartheta}$ parameter and use the lag structure as well as a time series identifier as a categorical effect in the additive predictor of $\beta$. We further investigate different number of BSPs $M\in\{5,10,30\}$ and different number of lags $p \in \{1,2,3\}$. Model training for all models but the ARIMA model was done using 1000 epochs with early stopping and a batch size of 128. For the MDN, we define 3 mixtures and use the AT($p$)'s $\beta$ as an additive predictor for the mean of every mixture component. The MVN uses the time series identifier to learn individual means and variances. 
For ARIMA we used the \texttt{auto.arima} implementation \citep{forecast} and performed a step-wise search via the AICc with different starting values for the order of the AR and the MA term. For the AR term, we consider the length of the corresponding forecasting horizon and halve this value. The search space for the MA term started either with 0 or 3. We chose the ARIMA model with the lowest AICc on the validation set. For the \texttt{auto.arima} model on the \texttt{m4} data, we restrict the observations to be used for model selection to 242 in order to reduce the computational complexity. A larger number did not give higher logscores. 


\paragraph{Datasets} We compare approaches on commonly used benchmark datasets electricity \citep[\texttt{elec;}][]{yu.2016}, traffic forecasting \citep[\texttt{traffic};][]{yu.2016}, monthly \texttt{tourism} \citep{Athanasopoulos.2011}, the hourly \texttt{m4} dataset \citep{Makridakis.2018} and currency \texttt{exchange} \citep{Lai.2018}. A short summary of these datasets can be found in Table~D3 in the Supplementary Material. 
\paragraph{Evaluation} For each proposed method and dataset, we report the log-scores \citep{gneiting.2007} and average results across time series and time points. The datasets are split into a training, validation, and test set by adhering to their time ordering. Evaluation windows are defined as done in the reference given for every dataset.

\begin{table*}[!ht]
    \centering
    \small
    \caption{Mean log-scores (higher is better) across 10 different initializations with standard deviations in brackets for each method (columns) and benchmark dataset (rows). Results for ARIMA are based on only one trial as there is typically no stochasticity in its results. The best performing method per data set is highlighted in bold.}
      \label{tab:crps_bench}
     \vspace*{0.3cm}
    \begin{tabular}{p{2.3cm}ccccc}
        ~ & ARIMA & AT($p$) & BoxCox & MDN & MVN \\ \hline
 \texttt{elec} & -5.44 & -5.35 (0.01) & -8.37 (0.00) & \textbf{-5.20} (0.01) & -9.51 (0.00) \\ 
\texttt{exchange} & 0.37 & \phantom{-}3.50 (0.05) & -0.69 (0.00) & \phantom{-}\textbf{4.02} (0.12) & -0.70 (0.00) \\ 
\texttt{m4} & -573.11& \textbf{-6.72} (0.07) & -10.7 (0.00) & -6.75 (1.17) & -12.0 (0.00) \\ 
\texttt{tourism} & -9.78 &  \textbf{-9.38} (0.01) & -11.5 (0.00) & -77.8 (99.5) & -12.7 (0.00) \\ 
\texttt{traffic}  & 0.23 & \phantom{-}\textbf{1.09} (0.33) & \phantom{-}0.03 (0.00) & \phantom{-}1.06 (0.02) & -0.25 (0.00) \\ 
\end{tabular}
\end{table*}
\paragraph{Results} Table~\ref{tab:crps_bench} shows the results of the comparison. Our approach always yields competitive and consistently good results while outperforming other models on most data sets. 


\section{Conclusion and Outlook} \label{sec:concl}

We have proposed ATMs, a flexible and comprehensible model class combining and extending various existing modeling approaches. ATMs allow for expressive probabilistic forecasts using a base distribution and a single transformation modeled by Bernstein polynomials. Additionally, a parametric inference paradigm based on MLE allows for statistical inference statements. ATMs 
empirically and theoretically recover well-known models, and demonstrate competitive performance on real-world datasets.

ATMs are the first adaption of transformation models to time series applications. Although our approach can be easily extended to incorporate deep neural network architectures, this invalidates statistical inference statements (e.g., because the uniqueness of $\bm{\theta}^*$ cannot be guaranteed). Future research will investigate this trade-off between larger model complexity and less statistical guarantees for the model. 

\backmatter

\bmhead{Acknowledgments}

DR has been partially supported by the German Federal Ministry of Education and Research (BMBF) under Grant No. 01IS18036A. TK gratefully acknowledges funding by the Deutsche Forschungsgemeinschaft (DFG, German Research Foundation), grant KN 922/9-1. TH was supported by the Swiss National Science Foundation, grant number 200021\_184603.

\section*{Declarations}

The authors declare that they have no known competing financial interests or personal relationships that could have appeared to influence the work reported in this paper.

\begin{appendices}




\section{Further Details} \label{app:theory}

\subsection{Definitions}

The following definition of the error distribution follows \citet{mlt.2018}.

\begin{definition}{\textbf{Error Distributions} \label{def:FZ}}
Let $Z:\Omega \to \mathbb{R}$ be a $\mathfrak{U}-\mathfrak{B}$ measurable function from $(\Omega, \mathfrak{U})$ to the Euclidian space with Borel $\sigma$-algebra $\mathfrak{B}$ with absolutely continuous distribution $\mathbb{P}_Z = f_Z \odot \mu_L$ on the probability space $(\mathbb{R}, \mathfrak{B},\mathbb{P}_Z)$ and $\mu_L$ the Lebesque measure. We define $F_Z$ and $F_Z^{-1}$ as the corresponding distributions and assume $F_Z(-\infty) = 0$, $F_Z(\infty) = 1$. $0 < f_Z(z) < \infty \,\, \forall z \in \mathbb{R}$ with log-concave, twice-differentiable density $f_Z$ with bounded first and second derivatives.
\end{definition}

\subsection{Propositions}

\begin{proposition}[\textbf{Interpretation of \eqref{eq:sats}}]\label{propo:sats} The ATM as defined in \eqref{eq:ATs} and further specified in \eqref{eq:sats} can be seen as an additive regression model with outcome $h_{1t}(y_t)$, predictor $h_{2t}((h_{1t} \odot \mathcal{Y}_{t}\mid  \mathcal{F}_{t-1}, \bm{x}) \mid  \bm{x})$ and error term $\varepsilon \sim F_Z.$
\end{proposition}

\proof{ 
We first define an additive regression model with outcome $\lambda_1 := h_{1t}(y_t)$, predictor $\tilde{\lambda}_2 := - h_{2t}((h_{1t} \odot \mathcal{Y}_{t}\mid  \mathcal{F}_{t-1}, \bm{x}) \mid  \bm{x})$ and error term $\varepsilon \sim F_Z$, i.e.,
$$\lambda_1 = \tilde{\lambda}_2 + \varepsilon, \, \varepsilon \sim F_Z,$$ where we use $\tilde{\lambda}_2 = -\lambda_2$ instead of $\lambda_2$ for convenience without loss of generality. This implies that ${\lambda}_1 - \tilde{\lambda}_2 = \lambda_1 + \lambda_2 = \varepsilon$ or equally ${\lambda}_1 + {\lambda}_2 \sim F_Z$. Optimizing this model is equal to fitting an ATM as defined in \eqref{eq:ATs} with structural assumption as defined in \eqref{eq:sats}.
}

\begin{proposition}[\textbf{Equivalence of AR($p$) and AT($p$) models}]\label{propo:arp} An autoregressive model of order $p$ (AR($p$)) with independent white noise following the distribution $F_Z$ in the location-scale family is equivalent to an AT($p$) model for $M=1$, $\bm{\vartheta}(\bm{x}) \equiv \bm{\vartheta}$, $r(\bm{x})\equiv 0$ and error distribution $F_Z$.
\end{proposition}

\proof{ 
The transformation function of an AT($p$) model with BSPs of order $M$ defined on an interval $[\iota_l, \iota_u]$, $\bm{\vartheta}(\bm{x}) \equiv \bm{\vartheta}$ and $r(\bm{x})\equiv 0$ is given by
\begin{equation*}
    \begin{split}
h_{1t} + h_{2t} = \bm{a}(y_t)^\top \bm{\vartheta} + \sum_{j=1}^p \phi_j \bm{a}(y_{t-j})^\top \bm{\vartheta}.
    \end{split}
\end{equation*}
We can further simplify the model by making $\bm{a}(y_t)$ more explicit: 
\begin{equation*}
    \begin{split}
\bm{a}(y_t) = (M+1)^{-1} \begin{pmatrix} f_{BE(1,M+1)}(\tilde{y}_t)\\ \vdots\\ f_{BE(m,M-m+1)}(\tilde{y}_t)\\ \vdots\\ f_{BE(M+1,1)}(\tilde{y}_t)\end{pmatrix} \in \mathbb{R}^{M+1} 
    \end{split}
\end{equation*}
with $\tilde{y}_t = (y-\iota_l)/(\iota_u-\iota_l)$ and Beta distribution density $f_{BE(\kappa,\mu)}$ with parameters $\kappa,\mu$. For simplicity and w.l.o.g. assume that $y_t \equiv \tilde{y}_t$. Setting $M$ to $1$, we get 
\begin{equation*}
    \begin{split}
        h_{1t} &= (\vartheta_0 f_{BE(1,2)} +  \vartheta_1 f_{BE(2,1)})/2\\ 
        &= \vartheta_0(1-{y}_t) + \vartheta_1{y}_t\\ 
        &= \vartheta_0 + (\vartheta_1-\vartheta_0) {y}_t\\ 
        &= \vartheta_0 + \tilde{\vartheta}_1 y_t.
    \end{split}
\end{equation*}
The transformation of the AT($p$) model is thus given by
\begin{equation} \label{eq:lemma1_1}
\begin{split}
h_t(y_t \mid  \mathcal{F}_{t-1}, \bm{x}) &= \vartheta_0 + \tilde{\vartheta}_1 y_t + \sum_{j=1}^p \phi_j (\vartheta_0 + \vartheta_1 y_{t-j})\\ &= \frac{y_t + \tilde{\vartheta}_0 + \sum_{j=1}^p \tilde{\phi}_j y_{t-j}}{\tilde{\vartheta}_1^{-1}} 
\end{split}
\end{equation}
with $\tilde{\vartheta}_0 = (\vartheta_0 (1 + \sum_j \phi_j))/\tilde{\vartheta_1}$ and $\tilde{\phi}_j = \phi_j \vartheta_1 / \tilde{\vartheta_1}$.
From \eqref{eq:ATs} we know
\begin{equation}\label{eq:lemma1_2}
    \mathbb{P}(Y_t \leq y_t \mid  \mathcal{F}_{t-1}, \bm{x}) = F_Z(h_t(y_t\mid  \mathcal{F}_{t-1}, \bm{x})).
\end{equation}
The AR($p$) model with coefficients $\varphi_0, \ldots, \varphi_p$ is given by
\begin{equation}\label{eq:lemma1_3}
\begin{split}
& y_t = \varphi_0 + \sum_{j=1}^p \varphi_j y_{t-j} + \sigma \varepsilon_t,\, \varepsilon_t \sim F_Z \\ &\Leftrightarrow \quad {Z} = \frac{y_t - \varphi_0 - \sum_{j=1}^p \varphi_j y_{t-j}}{\sigma} \sim F_Z.
\end{split}
\end{equation}
The equivalence of \eqref{eq:lemma1_2} in combination \eqref{eq:lemma1_1} with \eqref{eq:lemma1_3} is then given when setting $\tilde{\vartheta}_0 = - \varphi_0$, $\tilde{\phi}_j = -\varphi_j \forall j\in\{1,\ldots,p\}$ and $\sigma = \tilde{\vartheta}^{-1}_1$. Since both models find their parameters using Maximum Likelihood and it holds $\tilde{\vartheta}_1 > 0$ (as required for $\sigma$) by the monotonicity restriction on the BSPs coefficient, the models are identical up to different parameterization.
}

\subsection{Proof of Theorems}

The provided theorems~\ref{theorem:cons}-\ref{theorem:est} can be proven by observing that AT($p$)s' model structure and all made assumptions follow the general asymptotic theory for time series models as given in \citet{Ling.2010}. It is left to show that our setup and assumptions are equivalent to this general theory. 

\emph{Proof}. Our setup described in Section~\ref{sec:atp} together with Assumption~\ref{assum:th1}(i) corresponds to the setup described in \citet{Ling.2010}, Section 2. Our Assumption~\ref{assum:th1}(ii-iv) corresponds to their Assumption 2.1. In contrast, we do not consider the case of infinite $\mathcal{Y}_0$, but the extension is straightforward, by replacing initial values by some constant. Since AT(p)s and non-linear extensions are fully-parameterized time series models (Equation~\ref{eq:sats}) with parameter estimator $\hat{\bm{\theta}}_T$ found by MLE, all necessary assumptions are met to apply Theorem 2.1 in \citet{Ling.2010} including the subsequent remark, which yields the proof of our theorems \ref{theorem:cons}-\ref{theorem:est}.\qed

\section{Interpretability Example} \label{app:interp}

Next to the theoretical properties of ATMs described in Section~\ref{sec:subclass}, we will give an illustrative example in this section to make the different interpretability aspects of ATMs more tangible.

\begin{example}
Assume that the true generating process is additive on a log-scale and influenced by the two previous time points $t-1$ and $t-2$. For example, $t$ can be thought of as days in a year and the process $Y_t$ is an interest rate. Assume that the interest rate is multiplicatively influenced by the year $x_t \in E$ and further differs in its mean depending on a cyclic effect of the month $\eta_t$. An example for a corresponding data generating process would be 
\begin{equation*}
\begin{split}
\log(y_t) = &0.5\log(y_{t-1})\left(\sum_{e\in E}\theta_e I(x_t=e)\right) + \\ &0.2\log(y_{t-2})\left(\sum_{e\in E}\theta_e I(x_t=e)\right) +\\ 
&\sin(\eta_t) + \varepsilon_t,\quad \varepsilon_t \sim F_Z.    
\end{split}
\end{equation*}
In this case, the transformation function $h_{1t}$ can be defined as $h_{1t}(y_t) = \log(y_t)(\sum_{e\in E}\theta_e I(x_t=e))$ and approximated by $\bm{a}(y_t)^\top\bm{\vartheta}(x_t)$, where $\bm{a}$ is the BSP evaluation of $y_t$ and $\bm{\vartheta}$ a vector of coefficients depending on the year $x_t$. Further $\phi_1 = 0.5, \phi_2 = 0.2$, and the exogenous shift $r = \sin(\eta_t)$, which in practice would be approximated using a basis function representation. The interpretability properties listed in Section~\ref{sec:subclass} can be explained as follows: 
\begin{enumerate}
\item The additivity assumption in $\bm{\vartheta}$ allows to interpret the individual effects of the year $x_t$ on the transformation function $h_1$ individually (ceteris paribus) as $\log(y_t)(\sum_{e\in E}\theta_e I(x_t=e)) = \sum_{e\in E} \log(y_t)\theta_e I(x_t=e)$. Here, this would allow statements how a certain year $e$ influences the interest rate's density.
\item The use of the BSP basis for $\bm{a}$ in combination with 1. allows to visualize a forecasted density analytically for every additive term in $\bm{\vartheta}$. For example, to interpret year $e$, we evaluate $\bm{\vartheta}(x_t = e)$ and visualize $h_{1t}(y) = \bm{a}(y)^\top \bm{\vartheta}(x_t = e)$ as a function of $y$ on a given domain of interest. 
\item The structural assumptions of ATMs, i.e., their separation into two transformation functions $h_1$ and $h_2$, allows to interpret both transformation functions $h_1, h_2$ individually (ceteris paribus). In this example, the effect of the year can be interpreted using 1. and 2. while keeping the month fixed, and vice versa, the effect of the month can be interpreted by fixing the year. The applied transformation $h_1$ for AT($p$) models further allows to to individually interpret the influence of different lags (here these are the multiplicative effects $phi_1 = 0.5$ and $\phi_2=0.2$). 
\end{enumerate}
\end{example}

\section{Parametric Bootstrap} \label{app:sample}

To assess the parameter uncertainty included in the estimated density, we propose to use a parametric Bootstrap \citep[similar to the one suggested in][]{mlt.2018} that is based on the following steps:
\begin{enumerate}
    \item Generate $\hat{\bm{\theta}}^{(\nu)}, \nu=1,\ldots,N$ from the limiting distribution (Theorem 2 and 3);
    \item Draw samples $Z_{\tilde{t}} \sim F_Z, \tilde{t} \in \mathcal{T}$ and calculate $Y_{\tilde{t},\nu} = \inf \{y \in \Xi \mid  h_{\tilde{t}}(y,\hat{\bm{\theta}}^{(\nu)}) \geq Z_{\tilde{t}}\}$;
    \item Refit the model for each data set $\{Y_{\tilde{t},\nu}\}_{\tilde{t}\in\mathcal{T}}, \nu=1,\ldots,N$;
    \item Calculate the $N$ model densities.
\end{enumerate} 
Based on these $N$ model densities, uncertainty in the originally estimated density can be analyzed, e.g., visually by plotting all densities together as done in Figure~\ref{fig:motivating} and~\ref{fig:aleaepi}.

\section{Experimental Setup} \label{app:exp}

\subsection{Simulations}

In this subsection, we describe the details of the data generating process used in Figure~\ref{fig:motivating} (Section~\ref{sec:toy}) and provide results on experiments for the \emph{equivalence and consistency} paragraph of Section~\ref{sec:exp_arp} in Section~\ref{app:arp}.

\subsubsection{Data Generating Process Toy Example} \label{sec:toy}

For Figure~\ref{fig:motivating} we simulate $T=1000$ time points $y_1,\ldots,y_T$  that exhibit two modes as follows:
\begin{enumerate}
    \item Set $y_0 = 0$;
    \item Define a shift $\varrho=2$ and sample $x_1,\ldots,x_T$ from $\{-\varrho, \varrho\}$ with equal probability;
    \item Define a autoregressive coefficient $\phi_1=0.1$
    \item For $t=1,\ldots,T$, sample $y_t \sim \mathcal{N}(\phi_1 y_{t-1} + x_t, 1)$
\end{enumerate}
When providing the model with the marginal distribution of $y_t$ and defining $x_t$ as latent, unobserved variable, $y_t$ will exhibit two modes centered around $\pm \varrho$.

\subsubsection{AR($p$) comparison} \label{app:arp}

The data generating process for the simulation of Section~\ref{sec:exp_arp} is an AR model with the $p$ first coefficients 0.4, 0.2, 0.1, 0.05, 0.025. A standard implementation for the AR model was used. For the AT model we use the implementation provided in \citet{Ruegamer.2021} using 2500 epochs, batch size of 50, and early stopping based on 10\% of the training data. 

\begin{table}[]
    \caption{Average MSE in percent (with standard deviation in brackets) of estimated coefficients by the AR($p$) and AT($p$) model (rows) for different simulation settings (columns) over 100 replications.}
    \label{tab:arp}
    \centering
    \footnotesize
\begin{tabular}{cc|ccc}
 & &  $p=1$ & $p=2$ & $p=5$ \\ \hline
\multirow{2}{*}{$T=200$} & AR($p$) & 0.54 (0.73) & 0.49 (0.49) & 0.55 (0.4) \\ 
&  AT($p$) & 0.73 (1) & 0.68 (0.6) & 0.69 (0.42) \\ \hline
\multirow{2}{*}{$T=1000$} &  AR($p$)  & 0.12 (0.16) & 0.12 (0.13) & 0.12 (0.09) \\
& AT($p$)  & 0.17 (0.25) & 0.15 (0.16) & 0.17 (0.11) \\ \hline
\multirow{2}{*}{$T=5000$} & AR($p$) & 0.019 (0.03) & 0.02 (0.02) & 0.02 (0.02)\\
& AT($p$)  & 0.06 (0.09) & 0.05 (0.05) & 0.05 (0.03) 
\end{tabular}
\end{table}

\begin{table}[ht]
\centering
\footnotesize
\caption{Mean and standard deviation (brackets) of the mean squared error ($\times 10^2$ for better readability) between estimated and true coefficients in an AR($p$) model using our approach on the tampered data (bottom row) and the corresponding oracle based on the true data (Oracle).}
\label{tab:sats2}
\begin{tabular}{ccccc}
 &  &  $p=1$ & $p=2$ & $p=4$ \\ \hline
\multirow{2}{*}{$T=200$} &  Oracle & 0.65 (0.84) &  0.45 (0.46) & 0.46 (0.32) \\ 
&  AT($p$) & 0.49 (0.62) &  0.57 (0.76) & 0.65 (0.45) \\ \hline
\multirow{2}{*}{$T=400$} &  Oracle & 0.33 (0.31) & 0.22 (0.19) & 0.25 (0.13)  \\ 
&  AT($p$) & 0.52 (0.46) & 0.33 (0.3) & 0.34 (0.23) \\ \hline
\multirow{2}{*}{$T=800$} &  Oracle & 0.27 (0.34) & 0.13 (0.12) & 0.13 (0.085) \\ 
&  AT($p$) & 0.26 (0.36) & 0.17 (0.17) & 0.18 (0.12) \\ \hline
\end{tabular}
\end{table}

\subsection{Details on the benchmark study}\label{sec:SuppBench}

\subsubsection{Datasets}

Table~\ref{tab:datasets} summarizes the characteristics of the data sets used. For \texttt{elec} and \texttt{traffic} we use the 24 hours forecasting horizon and a pre-defined subset of one week of data. For \texttt{m4} and \texttt{tour} the test sets are already pre-defined with 48 hours and 24 months forecast windows, respectively. 

\begin{table*}[h]
    \centering
    \caption{Characteristics of the benchmark datasets.}
    \label{tab:datasets}
    \begin{tabular}{c|c|c|c|c|c}
         & \texttt{electricity} & \texttt{exchange} & \texttt{traffic} & \texttt{tourism} & \texttt{m4}\\
         \hline
        \# time series & 370 & 8 & 963 & 366 & 414 \\
        frequency & hourly & daily & hourly &  monthly & hourly \\
        forecast horizon & 24/72 & 1219 & 24/72 & 24 & 48 \\
        \# training samples & 71040 & 39048  & 184896 & 10980 & 269514 \\
    \end{tabular}
\end{table*}

\paragraph{Electricity} The dataset is available at \url{https://archive.ics.uci.edu/ml/datasets/ElectricityLoadDiagrams20112014}. According to \citet{Chen.2020}, Appendix A.3, the dataset describes the series of the electricity consumption (kWh) of 370 customers. The electricity usage values are recorded per 15 minutes from 2011 to 2014. We select the data of the last three years. By aggregating the records of the same hour, we use the hourly consumption data of size $370 \cdot 26304$, where 26304 is the length of the time series \citep{yu.2016}. The data used for modelling ranges from '2014-06-07 23:00:00' to '2014-06-09 23:00:00' including 1 day of validation and test data.

\paragraph{Exchange} The dataset is available from \citet{Lai.2018} and contains 8 bilateral exchange rate series for business days between Jan 1991 and May 2013. The split between training (60\%), validation (20\%) and test (20\%) is done based on the chronological order.

\paragraph{Traffic} The traffic dataset is available at \url{https://archive.ics.uci.edu/ml/datasets/PEMS-SF}. It describes the occupancy rates (between 0 and 1) of 963 car lanes of San Francisco bay area freeways. The measurements are carried out over the period from 2008-01-01 to 2009-03-30 and are sampled every 10 minutes. The original dataset is split into training and test. Hourly aggregation is applied to obtain hourly traffic data \citep{yu.2016}. The final time series are of length 10560 (the occupancy rates). The data used for modelling ranges from '2008-05-01 00:00:00' to '2008-05-09 23:00:00' including 1 day of validation and test data.

\paragraph{Tourism} The dataset is available at \url{https://robjhyndman.com/publications/the-tourism-forecasting-competition/}. Data is available on a monthly, quarterly and yearly level. We used the 366 monthly series which measure tourism demand. The data is split into test and train. 67 month are the minimum that is available for training and forecasting horizon is defined to be 24 months. The starting date for each monthly series is different. See Section 4 of \citet{Athanasopoulos.2011} for details.

\paragraph{m4} The dataset is taken from \citet{Makridakis.2018}. It contains 414 time series which are summarized in the m4 hourly data set. The split between training and test is already provided. Details on further background can be found on Wikipedia: \url{https://en.wikipedia.org/wiki/Makridakis_Competitions}. The starting point of each series is different. The minimum training length is 700 hours. The forecasting horizon is 48 hours.

\paragraph{Software}

For ATMs we extended the software \texttt{deepregression} \citep{Ruegamer.2021} by including an additional additive component for lags and used optimization techniques considered in \citet{ruegamer.nmdr.2020, Baumann.2020}. For ARIMA, we use the \texttt{forecast} R package \citep{forecast}.

\subsubsection{Computational Setup}


    All models were run on a server with 90GB RAM, 20 vCPUs from type Intel Xeon Processor (Skylake, IBRS), and a server with 64GB RAM, 32 vCPUs from type Intel(R) Xeon(R) CPU E5-2650 v2 @ 2.60GHz.

\section{Run-time Complexity}

In addition to forecasting performance comparisons, we also conduct a run-time benchmark to compare the run-time complexity of ATMs with other approaches. We use two different implementations for ATMs and measure their run-time. We contrast these run-times with the ARIMA model as implemented in the \texttt{forecast} R package \citep{forecast} and additionally include Prophet from the \texttt{prophet} R package \citep{prophet} as another fast alternative method for Bayesian forecasting.

The timing benchmark results (averaged over 10 replications) for different numbers of observations $T$ are given in Table~\ref{tab:runtime}. 
\begin{table}[!ht]
\begin{footnotesize}
    \centering
    \caption{Comparison of run-times for different methods (in columns) on different numbers of observations (\#Obs.) $T$ (in rows).} \label{tab:runtime}
    \begin{tabular}{r|rrrr}
        \#Obs. & ATM (plain) & ARIMA & Prophet & ATM (neural) \\ \hline
        $10^2$ & 0.199 & 0.005 & 0.372 & 22.20 \\ 
        $10^3$ & 0.513 & 0.024 & 0.097 & 31.30 \\ 
        $10^4$ & 3.920 & 0.118 & 0.342 & 28.80 \\ 
        $10^5$ & 94.62 & 1.121 & 33.99 & 32.30 \\ 
    \end{tabular}
    \end{footnotesize}
\end{table}
Results suggest that - as expected – ATMs in a neural network are very slow compared to ARIMA, Prophet and also a plain ATM implementation in \texttt{R}. However, all methods show an exponential increase in time consumption while the time consumption of the neural network implementation of ATMs (ATM (neural)) with mini-batch training and early stopping does only slightly increase in runtime for an exponential increase in number of observations. Moreover, for $10^5$ observations, ATM (plain) and Prophet already yield longer runtimes. 




\end{appendices}


\bibliography{sn-bibliography}

\begin{thebibliography}{50}
\providecommand{\natexlab}[1]{#1}
\providecommand{\url}[1]{{#1}}
\providecommand{\urlprefix}{URL }
\providecommand{\doi}[1]{\url{https://doi.org/#1}}
\providecommand{\eprint}[2][]{\url{#2}}
 \bibcommenthead

\bibitem[{Athanasopoulos et~al(2011)Athanasopoulos, Hyndman, Song, and
  Wu}]{Athanasopoulos.2011}
Athanasopoulos G, Hyndman RJ, Song H, et~al (2011) The tourism forecasting
  competition. International Journal of Forecasting 27(3):822--844

\bibitem[{Baumann et~al(2021)Baumann, Hothorn, and R{\"u}gamer}]{Baumann.2020}
Baumann PFM, Hothorn T, R{\"u}gamer D (2021) {Deep Conditional Transformation
  Models}. In: Machine Learning and Knowledge Discovery in Databases. Research
  Track. Springer International Publishing, Cham, pp 3--18

\bibitem[{Bengio and Bengio(1999)}]{Bengio.1999}
Bengio Y, Bengio S (1999) Modeling high-dimensional discrete data with
  multi-layer neural networks. MIT Press, NIPS'99, p 400–406

\bibitem[{Bernstein(1912)}]{Bernstein.1912}
Bernstein S (1912) D\'{e}monstration du th\'{e}or\`{e}me de weierstrass
  fond\'{e}e sur le calcul des probabilit\'{e}s. Communications of the Kharkov
  Mathematical Society 13(1):1--2

\bibitem[{Bishop(1994)}]{Bishop.1994}
Bishop CM (1994) Mixture density networks

\bibitem[{Chen et~al(2020)Chen, Vaughan, Nair, and Sudjianto}]{Chen.2020}
Chen J, Vaughan J, Nair VN, et~al (2020) {Adaptive Explainable Neural Networks
  (AxNNs)}. arXiv preprint arXiv:200402353
  {\href{https://arxiv.org/abs/2004.02353}{{https://arxiv.org/abs/arXiv:2004.02353}}}

\bibitem[{Chernozhukov et~al(2013)Chernozhukov, Fern{\'a}ndez-Val, and
  Melly}]{Chernozhukov.2013}
Chernozhukov V, Fern{\'a}ndez-Val I, Melly B (2013) Inference on counterfactual
  distributions. Econometrica 81(6):2205--2268

\bibitem[{Cox(1972)}]{Cox.1972}
Cox DR (1972) Regression models and life-tables. Journal of the Royal
  Statistical Society: Series B (Methodological) 34(2):187--202

\bibitem[{Dinh et~al(2017)Dinh, Sohl-Dickstein, and Bengio}]{Dinh.2017}
Dinh L, Sohl-Dickstein J, Bengio S (2017) Density estimation using real nvp.
  In: 5th International Conference on Learning Representations, {ICLR} 2017,
  Toulon, France, April 24 - 26, 2017, Conference Track Proceedings

\bibitem[{Dunson(2010)}]{Dunson.2010}
Dunson DB (2010) Nonparametric {Bayes} applications to biostatistics. Bayesian
  nonparametrics 28:223--273

\bibitem[{Farouki(2012)}]{Farouki.2012}
Farouki RT (2012) The {Bernstein} polynomial basis: A centennial retrospective.
  Computer Aided Geometric Design 29(6):379--419

\bibitem[{Foresi and Peracchi(1995)}]{Foresi.1995}
Foresi S, Peracchi F (1995) The conditional distribution of excess returns: An
  empirical analysis. Journal of the American Statistical Association
  90(430):451--466

\bibitem[{Gneiting and Katzfuss(2014)}]{gneiting2014probabilistic}
Gneiting T, Katzfuss M (2014) Probabilistic forecasting. Annual Review of
  Statistics and Its Application 1:125--151

\bibitem[{Gneiting et~al(2007)Gneiting, Balabdaoui, and
  Raftery}]{gneiting.2007}
Gneiting T, Balabdaoui F, Raftery AE (2007) Probabilistic forecasts,
  calibration and sharpness. Journal of the Royal Statistical Society: Series B
  (Statistical Methodology) 69(2):243--268

\bibitem[{Granger and Andersen(1978)}]{granger1978invertibility}
Granger CW, Andersen A (1978) On the invertibility of time series models.
  Stochastic Processes and their Applications 8(1):87--92

\bibitem[{Hamilton(2010)}]{hamilton2010regime}
Hamilton JD (2010) Regime switching models. In: Macroeconometrics and time
  series analysis. Springer, p 202--209

\bibitem[{Hothorn(2020)}]{Hothorn.2020}
Hothorn T (2020) Transformation boosting machines. Statistics and Computing
  30(1):141--152

\bibitem[{Hothorn et~al(2014)Hothorn, Kneib, and B{\"u}hlmann}]{Hothorn.2014}
Hothorn T, Kneib T, B{\"u}hlmann P (2014) Conditional transformation models.
  Journal of the Royal Statistical Society: Series B: Statistical Methodology
  pp 3--27

\bibitem[{Hothorn et~al(2018)Hothorn, M{\"o}st, and B{\"u}hlmann}]{mlt.2018}
Hothorn T, M{\"o}st L, B{\"u}hlmann P (2018) Most likely transformations.
  Scandinavian Journal of Statistics 45(1):110--134

\bibitem[{Hyndman et~al(2021)Hyndman, Athanasopoulos, Bergmeir, Caceres, Chhay,
  O'Hara-Wild, Petropoulos, Razbash, Wang, and Yasmeen}]{forecast}
Hyndman R, Athanasopoulos G, Bergmeir C, et~al (2021) {forecast}: Forecasting
  functions for time series and linear models. R package version 8.15

\bibitem[{Jordan et~al(2002)}]{Jordan.2002}
Jordan A, et~al (2002) On discriminative vs. generative classifiers: A
  comparison of logistic regression and naive {Bayes}. Advances in neural
  information processing systems 14(2002):841

\bibitem[{Kastner et~al(2017)Kastner, Frühwirth-Schnatter, and
  Lopes}]{kastner2017}
Kastner G, Frühwirth-Schnatter S, Lopes HF (2017) Efficient bayesian inference
  for multivariate factor stochastic volatility models. Journal of
  Computational and Graphical Statistics 26(4):905--917

\bibitem[{Kingma and Ba(2014)}]{Kingma.2014}
Kingma DP, Ba J (2014) Adam: A method for stochastic optimization. arXiv
  preprint arXiv:14126980

\bibitem[{Kingma et~al(2016)Kingma, Salimans, Jozefowicz, Chen, Sutskever, and
  Welling}]{Kingma.2016}
Kingma DP, Salimans T, Jozefowicz R, et~al (2016) Improved variational
  inference with inverse autoregressive flow. In: Lee D, Sugiyama M, Luxburg U,
  et~al (eds) Advances in Neural Information Processing Systems, vol~29. Curran
  Associates, Inc.

\bibitem[{Koenker(2005)}]{Koenker.2005}
Koenker R (2005) Quantile Regression, vol Economic Society Monographs.
  Cambridge University Press

\bibitem[{Kook et~al(2021)Kook, Herzog, Hothorn, D{\"u}rr, and
  Sick}]{Kook.2020}
Kook L, Herzog L, Hothorn T, et~al (2021) Deep and interpretable regression
  models for ordinal outcomes. Pattern Recognition

\bibitem[{Kook et~al(2022)Kook, Götschi, Baumann, Hothorn, and
  Sick}]{Kook.2022}
Kook L, Götschi A, Baumann PF, et~al (2022) Deep interpretable ensembles.
  \urlprefix\url{https://arxiv.org/abs/2205.12729}

\bibitem[{Lai et~al(2018)Lai, Chang, Yang, and Liu}]{Lai.2018}
Lai G, Chang WC, Yang Y, et~al (2018) Modeling long-and short-term temporal
  patterns with deep neural networks. In: The 41st International ACM SIGIR
  Conference on Research \& Development in Information Retrieval, pp 95--104

\bibitem[{Lin et~al(1996)Lin, Horne, Tino, and Giles}]{Lin.1996}
Lin T, Horne BG, Tino P, et~al (1996) Learning long-term dependencies in narx
  recurrent neural networks. IEEE Transactions on Neural Networks
  7(6):1329--1338

\bibitem[{Ling and McAleer(2010)}]{Ling.2010}
Ling S, McAleer M (2010) A general asymptotic theory for time-series models.
  Statistica Neerlandica 64(1):97--111

\bibitem[{Liu et~al(2019)Liu, Paisley, Kioumourtzoglou, and Coull}]{Liu.2019}
Liu J, Paisley J, Kioumourtzoglou MA, et~al (2019) Accurate uncertainty
  estimation and decomposition in ensemble learning. In: Wallach H, Larochelle
  H, Beygelzimer A, et~al (eds) Advances in Neural Information Processing
  Systems, vol~32. Curran Associates, Inc.

\bibitem[{Makridakis et~al(2018)Makridakis, Spiliotis, and
  Assimakopoulos}]{Makridakis.2018}
Makridakis S, Spiliotis E, Assimakopoulos V (2018) The m4 competition: Results,
  findings, conclusion and way forward. International Journal of Forecasting
  34(4):802--808

\bibitem[{Murphy(2012)}]{Murphy.2012}
Murphy KP (2012) Machine learning: a probabilistic perspective. MIT press

\bibitem[{Papamakarios et~al(2017)Papamakarios, Pavlakou, and
  Murray}]{Papamakarios.2017}
Papamakarios G, Pavlakou T, Murray I (2017) Masked autoregressive flow for
  density estimation. In: Guyon I, Luxburg UV, Bengio S, et~al (eds) Advances
  in Neural Information Processing Systems

\bibitem[{Papamakarios et~al(2021)Papamakarios, Nalisnick, Rezende, Mohamed,
  and Lakshminarayanan}]{Papamakarios.2019}
Papamakarios G, Nalisnick E, Rezende DJ, et~al (2021) Normalizing flows for
  probabilistic modeling and inference. Journal of Machine Learning Research
  22(57):1--64

\bibitem[{Raftery et~al(2005)Raftery, Gneiting, Balabdaoui, and
  Polakowski}]{raftery2005using}
Raftery AE, Gneiting T, Balabdaoui F, et~al (2005) Using bayesian model
  averaging to calibrate forecast ensembles. Monthly weather review
  133(5):1155--1174

\bibitem[{Rao(1981)}]{rao1981theory}
Rao TS (1981) On the theory of bilinear time series models. Journal of the
  Royal Statistical Society: Series B (Methodological) 43(2):244--255

\bibitem[{R{\"u}gamer et~al(2020)R{\"u}gamer, Kolb, and Klein}]{Ruegamer.2020}
R{\"u}gamer D, Kolb C, Klein N (2020) {Semi-Structured Deep Distributional
  Regression: A Combination of Additive Models and Deep Learning}. arXiv
  preprint arXiv:200205777
  {\href{https://arxiv.org/abs/2002.05777}{{https://arxiv.org/abs/arXiv:2002.05777}}}

\bibitem[{R\"ugamer et~al(2020)R\"ugamer, Pfisterer, and
  Bischl}]{ruegamer.nmdr.2020}
R\"ugamer D, Pfisterer F, Bischl B (2020) Neural mixture distributional
  regression. arXiv preprint arXiv:201006889
  {\href{https://arxiv.org/abs/2010.06889}{{https://arxiv.org/abs/arXiv:2010.06889}}}

\bibitem[{{R{\"u}gamer} et~al(2022){R{\"u}gamer}, Kolb, Fritz, Pfisterer,
  Kopper, Bischl, Shen, Bukas, de~Andrade~e Sousa, Thalmeier, Baumann, Kook,
  Klein, and M{\"u}ller}]{Ruegamer.2021}
{R{\"u}gamer} D, Kolb C, Fritz C, et~al (2022) deepregression: a flexible
  neural network framework for semi-structured deep distributional regression.
  Journal of Statistical Software Accepted,
  {\href{https://arxiv.org/abs/2104.02705}{{https://arxiv.org/abs/arXiv:2104.02705}}}

\bibitem[{Sakia(1992)}]{Sakia.1992}
Sakia RM (1992) The box-cox transformation technique: a review. Journal of the
  Royal Statistical Society: Series D (The Statistician) 41(2):169--178

\bibitem[{Schlosser et~al(2019)Schlosser, Hothorn, Stauffer, and
  Zeileis}]{Schlosser_2019}
Schlosser L, Hothorn T, Stauffer R, et~al (2019) Distributional regression
  forests for probabilistic precipitation forecasting in complex terrain. The
  Annals of Applied Statistics 13(3)

\bibitem[{Shumway et~al(2000)Shumway, Stoffer, and Stoffer}]{Shumway.2020}
Shumway RH, Stoffer DS, Stoffer DS (2000) Time series analysis and its
  applications, vol~3. Springer

\bibitem[{Sick et~al(2021)Sick, Hothorn, and D{\"u}rr}]{Sick.2020}
Sick B, Hothorn T, D{\"u}rr O (2021) Deep transformation models: Tackling
  complex regression problems with neural network based transformation models.
  In: 2020 25th International Conference on Pattern Recognition (ICPR), IEEE,
  pp 2476--2481

\bibitem[{Taylor and Letham(2021)}]{prophet}
Taylor S, Letham B (2021) prophet: Automatic Forecasting Procedure. R package
  version 1.0

\bibitem[{Uria et~al(2016)Uria, C{{\^o}}t{{\'e}}, Gregor, Murray, and
  Larochelle}]{Uria.2016}
Uria B, C{{\^o}}t{{\'e}} MA, Gregor K, et~al (2016) Neural autoregressive
  distribution estimation. Journal of Machine Learning Research 17(205):1--37

\bibitem[{Van~Belle et~al(2011)Van~Belle, Pelckmans, Suykens, and
  Van~Huffel}]{Van.2011}
Van~Belle V, Pelckmans K, Suykens JA, et~al (2011) Learning transformation
  models for ranking and survival analysis. Journal of machine learning
  research 12(3)

\bibitem[{Wong and Li(2000)}]{Wong.2000}
Wong CS, Li WK (2000) On a mixture autoregressive model. Journal of the Royal
  Statistical Society: Series B (Statistical Methodology) 62(1):95--115

\bibitem[{Wu and Tian(2013)}]{Wu.2013}
Wu CO, Tian X (2013) Nonparametric estimation of conditional distributions and
  rank-tracking probabilities with time-varying transformation models in
  longitudinal studies. Journal of the American Statistical Association
  108(503):971--982

\bibitem[{Yu et~al(2016)Yu, Rao, and Dhillon}]{yu.2016}
Yu HF, Rao N, Dhillon IS (2016) Temporal regularized matrix factorization for
  high-dimensional time series prediction. In: NIPS, pp 847--855

\end{thebibliography}


\end{document}